\documentclass[10pt,twocolumn,letterpaper]{article}
\usepackage[accsupp]{axessibility} 
\usepackage[pagenumbers]{cvpr} 

\usepackage{graphicx}
\usepackage{amsmath}
\usepackage{amssymb}
\usepackage{booktabs}
\usepackage{color}
\usepackage{diagbox}
\usepackage{microtype}
\usepackage[table, dvipsnames]{xcolor}
\usepackage{colortbl}
\usepackage{makecell}
\usepackage{amssymb}
\usepackage{enumitem}

\newcommand{\authorskip}{\hspace{12mm}}

\makeatletter

\newcommand{\Rmnum}[1]{\textcolor{red}{\expandafter\@slowromancap\romannumeral #1@}}

\makeatother

\usepackage[pagebackref,breaklinks,colorlinks]{hyperref}

\usepackage[capitalize]{cleveref}
\crefname{section}{Sec.}{Secs.}
\Crefname{section}{Section}{Sections}
\Crefname{table}{Table}{Tables}
\crefname{table}{Tab.}{Tabs.}

\def\cvprPaperID{6034} 
\def\confName{CVPR}
\def\confYear{2023}

\begin{document}

\title{High Fidelity 3D Hand Shape Reconstruction\\ via Scalable Graph Frequency Decomposition}

\author{
 Tianyu Luan$^{1}$ \authorskip Yuanhao Zhai$^{1}$ \authorskip Jingjing Meng$^{1}$ \authorskip
 Zhong Li$^{2}$ \\ Zhang Chen$^{2}$ \authorskip Yi Xu$^{2}$ \authorskip Junsong Yuan$^{1}$\\[3mm]
 $^1$State University of New York at Buffalo ~~~~~~ $^2$OPPO US Research Center, InnoPeak Technology, Inc. \\
 {\tt\small \{tianyulu,yzhai6,jmeng2,jsyuan\}@buffalo.edu}\\ {\tt\small\{zhong.li,zhang.chen,yi.xu\}@oppo.com}
}

\maketitle

\begin{abstract}
Despite the impressive performance obtained by recent single-image hand modeling
techniques, they lack the capability to capture sufficient details of the 3D
hand mesh.
This deficiency greatly limits their applications when high-fidelity hand
modeling is required, \eg{}, personalized hand modeling.
To address this problem, we design a frequency split network to generate 3D hand
mesh using different frequency bands in a coarse-to-fine manner.
To capture high-frequency personalized details, we transform the 3D mesh into
the frequency domain, and propose a novel frequency decomposition loss to
supervise each frequency component.
By leveraging such a coarse-to-fine scheme, hand details that correspond to the
higher frequency domain can be preserved.
In addition, the proposed network is scalable, and can stop the inference at any
resolution level to accommodate different hardware with varying computational powers.
To quantitatively evaluate the performance of our method in terms of recovering
personalized shape details, we introduce a new evaluation metric named Mean
Signal-to-Noise Ratio (MSNR) to measure the signal-to-noise ratio of each mesh
frequency component.
Extensive experiments demonstrate that our approach generates fine-grained
details for high-fidelity 3D hand reconstruction, and our evaluation metric is
more effective for measuring mesh details compared with traditional metrics. The code is available at \url{https://github.com/tyluann/FreqHand}.
\end{abstract}

\section{Introduction}

High-fidelity and personalized 3D hand modeling have seen great demand in 3D games, virtual reality, and the emerging Metaverse, as it brings better user experiences, \eg{}, users can see their own realistic hands in the virtual space instead of the standard avatar hands. Therefore, it is of great importance to reconstruct high-fidelity hand meshes that can adapt to different users and application scenarios.

\begin{figure}[t]
  \centering
  \includegraphics[width=\linewidth]{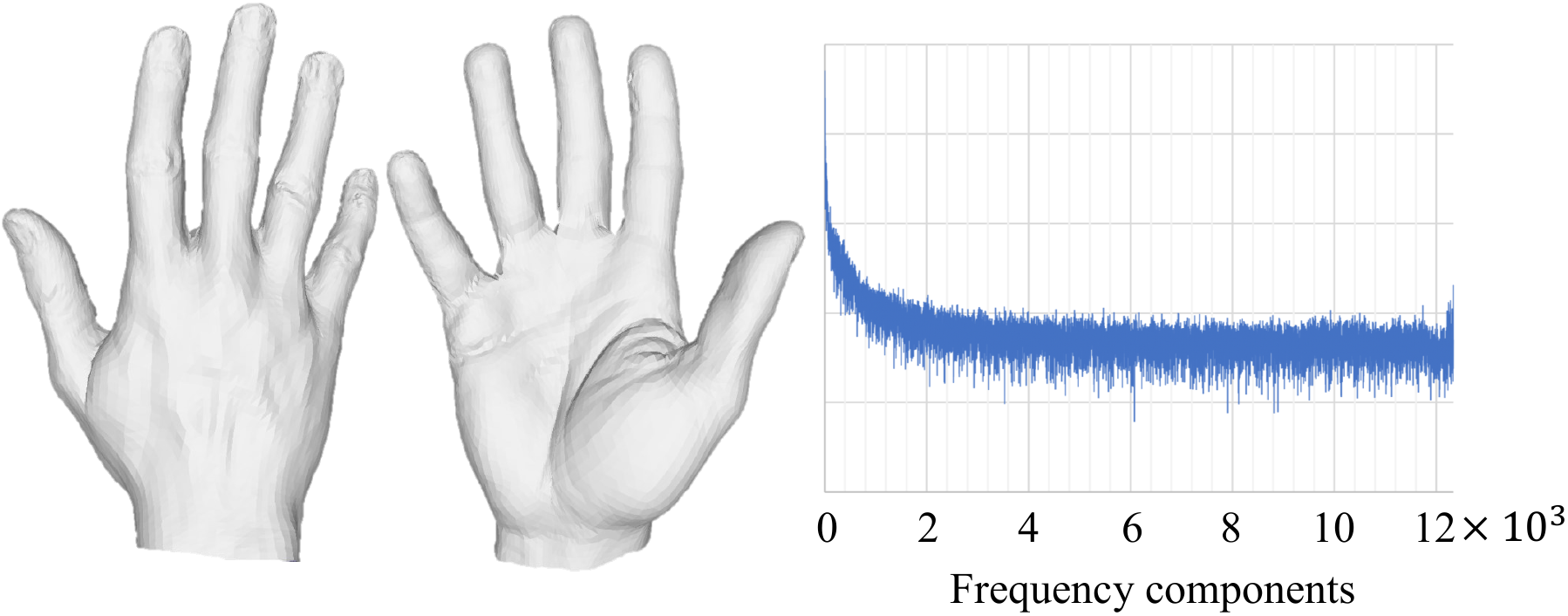}
  \caption{An exemplar hand mesh of sufficient details and its graph frequency decomposition. The x-axis shows frequency components from low to high. The y-axis shows the amplitude of each component in the logarithm. At the frequency domain, the signal amplitude generally decreases as the frequency increases.}
  \label{fig:spetrum}
\end{figure}

Despite previous successes in 3D hand reconstruction and modeling\cite{hasson2019learning,baek2020weakly,yang2020seqhand,chen2021mobrecon,yang2020bihand,tang2021towards,lin2021end,chen2021temporal}, few existing solutions focus
on enriching the details of the reconstructed shape, and most current methods fail to generate consumer-friendly high-fidelity hands.
When we treat the hand mesh as graph signals, like most natural signals, the low-frequency components have larger amplitudes than those of the high-frequency parts, which we can observe in a hand mesh spectrum curve (\cref{fig:spetrum}). Consequently, if we generate the mesh purely in the spatial domain, the signals of different frequencies could be biased, thus the high-frequency information can be easily overwhelmed by its low-frequency counterpart. 
Moreover, the wide usage of compact parametric models, such as MANO \cite{MANO:SIGGRAPHASIA:2017}, has limited the expressiveness of personalized details. Even though MANO can robustly estimate the hand pose and coarse shape, it sacrifices hand details for compactness and robustness in the parameterization process, so the detail expression ability of MANO is suppressed. 

To better model detailed 3D shape information, we transform the hand mesh into the graph frequency domain, and design a frequency-based loss function to generate high-fidelity hand mesh in a scalable manner. Supervision in the frequency domain explicitly constrains the signal of a given frequency band from being influenced by other frequency bands. Therefore, the high-frequency signals of hand shape will not be suppressed by low-frequency signals despite the amplitude disadvantage. 
To improve the expressiveness of hand models, we design a new hand model of $12,337$ vertices that extends previous parametric models such as MANO with nonparametric representation for residual adjustments. While the nonparametric residual expresses personalized details, the parametric base ensures the overall structure of the hand mesh, \eg{}, reliable estimation of hand pose and 3D shape. Instead of fixing the hand mesh resolution, we design our network architecture in a coarse-to-fine manner with three resolution levels U-net for scalability. Different levels of image features contribute to different levels of detail. Specifically, we use low-level features in high-frequency detail generation and high-level features in low-frequency detail generation. At each resolution level, our network outputs a hand mesh with the corresponding resolution. During inference, the network outputs an increasingly higher resolution mesh with more personalized details step-by-step, while the inference process can stop at any one of the three resolution levels.

In summary, our contributions include the following.
\begin{enumerate}[leftmargin=*,noitemsep,topsep=0pt]
    \item We design a high-fidelity 3D hand model for reconstructing 3D hand shapes from single images. The hand representation provides detailed expression, and our frequency decomposition loss helps to capture the personalized shape information.
    \item To enable computational efficiency, we propose a frequency split network architecture to generate high-fidelity hand mesh in a scalable manner with multiple levels of detail.
    During inference, our scalable framework supports budget-aware mesh reconstruction when the computational resources are limited. 
    \item We propose a new metric to evaluate 3D mesh details. It better captures the signal-to-noise ratio of all frequency bands to evaluate high-fidelity hand meshes. The effectiveness of this metric has been validated by extensive experiments.
\end{enumerate}

We evaluate our method on the InterHand2.6M dataset~\cite{Moon_2020_ECCV_InterHand2.6M}. In addition to the proposed evaluation metrics, we also evaluate mean per joint position error (MPJPE) and mesh Chamfer distance (CD). Compared to MANO and other baselines, our proposed method achieves better results using all three metrics.

\section{Related Work}
\textbf{Parametric hand shape reconstruction.} Parametric models are a popular approach in hand mesh reconstruction. Romero \etal{} \cite{MANO:SIGGRAPHASIA:2017} proposed MANO, which uses a set of shape and pose parameters to control the movement and deformation of human hands. Many recent works \cite{hasson2019learning,yang2020seqhand,tang2021towards,zhang2021hand,tu2022consistent,zhang2019end,zheng2021deformation,peng20203d} combined deep learning with MANO. They use features extracted from the RGB image as input, CNN to get the shape and pose parameters, and eventually these parameters to generate hand mesh. These methods make use of the strong prior knowledge provided by the hand parametric model, so that it is convenient to train the networks and the results are robust. However, the parametric method limits the mesh resolution and details of hands. 

\textbf{Non-parametric hand shape reconstruction.} Non-parametric hand shape reconstruction typically estimates the vertex positions of a template with fixed topology. For example, Ge \etal{}~\cite{ge20193d} proposed a method using a graph convolution network. It uses a predefined upsampling operation to build a multi-level spectrum GCN network. Kulon \etal{} \cite{kulon2020weakly} used spatial GCN and spiral convolution operator for mesh generation. Moon \etal{} \cite{moon2020i2l} proposed a pixel-based approach. However, none of these works paid close attention to detailed shapes. Moon \etal{} \cite{Moon_2020_ECCV_DeepHandMesh} provided an approach that outputs fine details, but since they need the 3D scanned meshes of the test cases for training, their model cannot do cross-identity reconstruction. 
In our paper, we design a new hand model that combines the strength of both parametric and non-parametric approaches. We use this hand model as a basis to reconstruct high-fidelity hands.

\textbf{Mesh frequency analysis.} Previous works mainly focused on the spectrum analysis of the entire mesh graph. Chung. \cite{chung1997spectral} defines the graph Fourier transformation and graph Laplacian operator, which builds the foundation of graph spectrum analysis. \cite{shuman2013emerging} extends commonly used signal processing operators to graph space. \cite{bruna2013spectral} proposes a spectrum graph convolution network based on graph spectrum characteristics. The spectral decomposition of the graph function is used to define graph-based convolution. Recent works such as \cite{kipf2016semi,schlichtkrull2018modeling,xing2019dynamic,shang2021multi,zhu2020simple,ma2019spectral,du2017topology} widely use spectrum GCN in different fields. However, these works mainly focus on the analysis of the overall graph spectrum. In this paper, we use spectrum analysis as a tool to design our provided loss function and metric.

\begin{figure*}[t]
  \centering
  \includegraphics[width=1\linewidth]{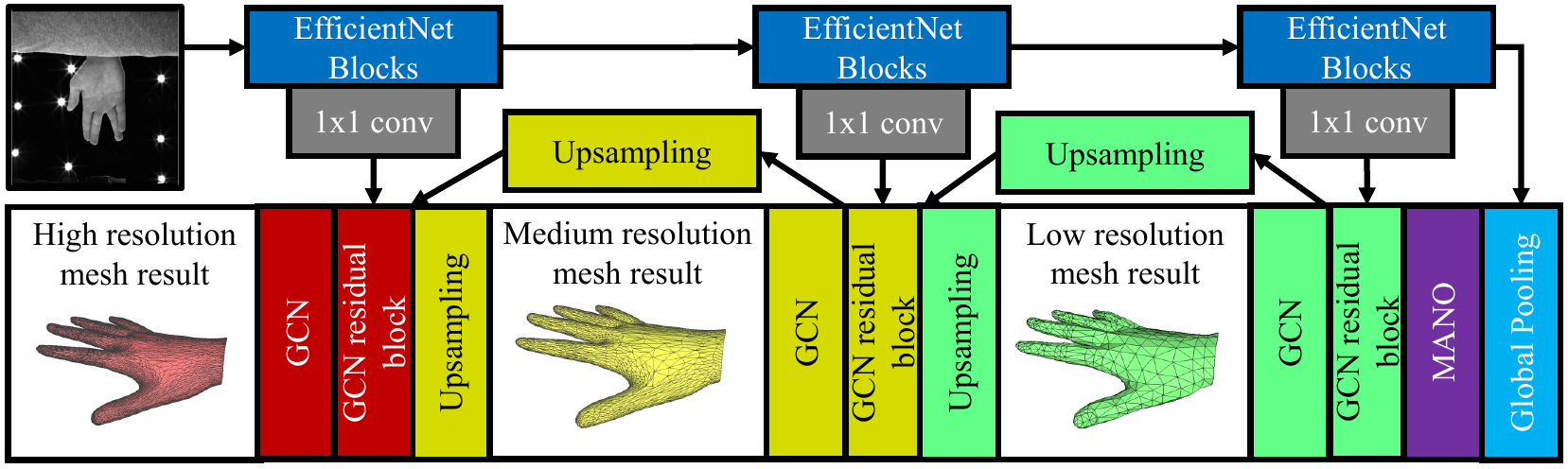}
  \caption{We design our scalable hand modeling network in a U-net manner. First, we generate a MANO mesh from image features (\textcolor{violet}{purple} block). Then, based on the MANO mesh, we use a multilevel GCN to recover 3 levels of personalized mesh (\textcolor{green}{green}, \textcolor{Goldenrod}{yellow}, and \textcolor{red}{red} blocks from low to high). In order to obtain high-frequency hand details, we use skip-connected image features from different layers of the backbone network (\textcolor{blue}{blue} and \textcolor{gray}{gray} blocks) 
  At inference, our network can stop at any resolution level, but still provides reasonable high-fidelity results at that resolution. The architecture and implementation details can be found in supplementary material Section \Rmnum{1} and \Rmnum{2}.}
  \label{fig:pipeline}
\end{figure*}

\section{Proposed Method}
We propose a scalable network that reconstructs the detailed hand shape, and use frequency decomposition loss to acquire details. \cref{fig:pipeline} shows our network architecture. We design our network in the manner of a U-net. First, we generate a MANO mesh from image features from EfficientNet~\cite{tan2019efficientnet}.
Based on the MANO mesh, we use a 
graph convolution network (green, yellow, and red modules in \cref{fig:pipeline}) to recover a high-fidelity hand mesh. In order to obtain high-frequency information, we use image features from different layers of the backbone network as a part of the GCN inputs. Specifically, at the low-resolution level, we take high-level image features as part of the input, and use a low-resolution graph topology to generate a low-resolution mesh. At medium and high-frequency levels, we use lower-level image feature through the skip connection to produce a high-resolution mesh. 
Note that at every resolution level, the network will output the intermediate hand mesh, so it would naturally have the ability for scalable inference. During the training process, we supervise both intermediate meshes and the final high-resolution mesh. We discuss the details in the following.

\subsection{High Fidelity 3D Hand Model}
\label{sec:rep}

We design our hand representation based on MANO~\cite{MANO:SIGGRAPHASIA:2017}. MANO factorizes human hands into a 10-dimensional shape representation $\beta$ and a 35-dimensional pose representation $\theta$.
MANO model can be represented as 
\begin{equation}
\left\{
\begin{aligned}
    &M(\theta, \beta) = W(T_{P}(\theta, \beta), \theta, w) \\
    &T_{P}(\theta, \beta) = \overline{T} + B_{S}(\beta) + B_{P}(\theta)
\end{aligned}
 \right.
\end{equation}
where $W$ is the linear blend skinning function. 
Model parameter $w$ is the blend weight. $B_{S}$ and $B_{P}$ are another two parameters of MANO named shape blend shape and pose blend shape, which are related to pose and shape parameters, respectively.
MANO can transfer complex hand surface estimation into a simple regression of a few pose and shape parameters. 
However, MANO has limited capability in modeling shape detail. 
It is not only limited by the number of pose and shape dimensions (45), but also by the number of vertices (778). In our work, we designed a new parametric-based model with 12,338 vertices generated from MANO via subdivision. The large vertex number greatly enhances the model's ability to represent details.

\textbf{Subdivided MANO.} To address this problem. We design an extended parametric model that can better represent details. First, we add detail residuals to MANO as
\begin{equation}
\begin{aligned}
    &M^{\prime}(\theta, \beta, d) = W(T_{P}^{\prime}(\theta, \beta, d), \theta, w^{\prime}), \\
    &T_{P}^{\prime}(\theta, \beta, d) = \overline{T}^{\prime} + B_{S}^{\prime}(\beta) + B_{P}^{\prime}(\theta) + d,
    \label{eq:mesh_rep1}
\end{aligned}
\end{equation}
where, $w^{\prime}$, $\overline{T}^{\prime}$, $B_{S}^{\prime}(\beta)$, and $B_{P}^{\prime}(\theta)$ are the parameters our model, and $d$ is the learnable per-vertex location perturbation. The dimension of $d$ is the same as the number of vertices.

Besides vertex residuals, we further increase the representation capability of our hand model by increasing the resolution of the mesh. 
Motivated by the traditional Loop subdivision\cite{loop1987smooth}, we propose to design our parametric hand model by subdividing the MANO template. Loop subdivision can be represented as 
\begin{equation}
    \overline{T}^{\prime} = \mathbf{L_s} \overline{T},
    \label{eq:mesh_rep2}
\end{equation}
where, $\overline{T}$ is original template mesh with $n$ vertices and $m$ edges. $\overline{T}^{\prime}$ is the subdivided template mesh with $n+m$ vertices. $\mathbf{L_s} \in \mathbb{R}^{(n+m)\times m}$ is the linear transformation that defines the subdivision process. The position of each vertex on the new mesh is only determined by the neighbor vertices on the original mesh, so $\mathbf{L_s}$ is sparse. We use similar strategies to calculate $B_{S}$ and $B_{P}$. The MANO parameters map the input shape and pose into vertex position adjustments. These mappings are linear matrices of dimension $x \times n$. 
Therefore, we can calculate the parameters as
\begin{equation}
\begin{aligned}
    w^{\prime} = (\mathbf{L_s}w^{\top})^{\top}, \\
    B_{S}^{\prime} = (\mathbf{L_s}B_{S}^{\top})^{\top}, \\
    B_{P}^{\prime} = (\mathbf{L_s}B_{P}^{\top})^{\top}.
    \label{eq:mesh_rep3}
\end{aligned}
\end{equation}
 We repeat the procedure twice to get sufficient resolution.

\begin{figure}[t]
  \centering
  \includegraphics[width=0.9\linewidth]{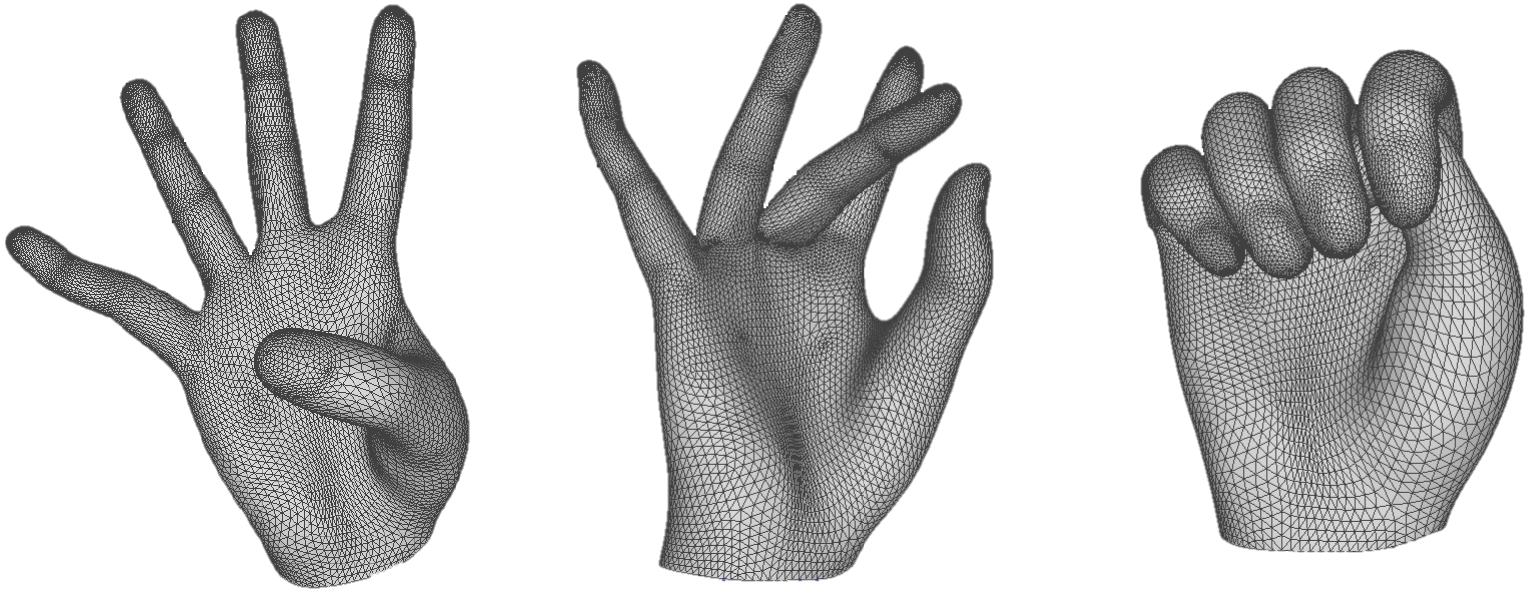}
  \caption{We design a new high-fidelity hand mesh with $12,337$ vertices. Our new model inherits the advantage of the parametric hand model and provides reliable 3D shape estimation with fewer flaws when hand poses change.}
  \label{fig:subdivision}
\end{figure}

 \cref{fig:subdivision} shows example meshes from the new model in different poses ($d$ is set to 0). We can see that our representation inherits the advantages of the parametric hand model. It has a plausible structure with no visual artifacts when the hand poses change. 

\subsection{Hierachical Graph Convolution Network}
\label{sec:gcn}
Our GCN network utilizes a multiresolution graph architecture 
that follows the subdivision process in Section \cref{sec:rep}.
Different from the single graph GCNs in previous works \cite{malik2021handvoxnet++,kong2020sia}, our GCN network uses different graphs in different layers. At each level, each vertex of the graph corresponds to a vertex on the mesh and the graph topology is defined by the mesh edges. Between two adjunct resolution levels, the network  uses the $\mathbf{L_s}$ in \cref{eq:mesh_rep2} for upsampling operation.

This architecture is designed for scalable inference. When the computing resources are limited, only the low-resolution mesh needs to be calculated; when the computing resources are sufficient, then we can calculate all the way to the high-resolution mesh. Moreover, this architecture allows us to explicitly supervise the intermediate results, so the details would be added level-by-level.

\subsection{Graph Frequency Decomposition}
\label{sec:freq}
In order to supervise the output mesh in the frequency domain and design the frequency-based metric, we need to do frequency decomposition on mesh shapes. Here, we regard the mesh as an undirected graph, and 3D locations of mesh vertices as signals on the graph. Then, the frequency decomposition of the mesh is the spectrum analysis of this graph signal. Following \cite{chung1997spectral}, given an undirected graph $\mathcal{G} = \left\{\mathcal{V}, \mathcal{E} \right\}$ with a vertices set of $\mathcal{V}= \left\{1,2,...,N \right\}$ and a set of edges $\mathcal{E}= \left\{(i, j) \right\}_{i,j \in \mathcal{V}}$, the Laplacian matrix is defined as $\mathbf{L}:=\mathbf{D} - \mathbf{A}$,
where $\mathbf{A}$ is the $N \times N$ adjacency matrix with entries defined as edge weights $a_{ij}$ and $\mathbf{D}$ is the diagonal degree matrix. The $i$th diagonal entry $d{i} = \sum_{j}a_{ij}$. In this paper, the edge weights are defined as
\begin{equation}
a_{ij}:=\left\{
\begin{aligned}
1 &, & (i,j) \in \mathcal{E} \\
0 &, & otherwise
\end{aligned}
 \right.
\end{equation}
which means all edges have the same weights. We decompose $\mathbf{L}$ using spectrum decomposition:
\begin{align}
\mathbf{L}=\mathbf{U}^{\top}\mathbf{\Lambda}\mathbf{U}.
\end{align}
Here, $\mathbf{\Lambda}$ is a diagonal matrix, in which the diagonal entries are the eigenvalues of $\mathbf{L}$. $\mathbf{U}$ is the eigenvector set of $\mathbf{L}$. Since the Laplacian matrix $\mathbf{L}$ describes the fluctuation of the graph signal, its eigenvalues show how "frequent" the fluctuations are in each eigenvector direction. Thus, the eigenvectors of larger eigenvalues are defined as higher frequency bases, and the eigenvectors of smaller eigenvalues are defined as lower frequency bases. Since the column vectors of $\mathbf{U}$ is a set of orthonormal basis of the graph space, following \cite{sardellitti2017graph}, we define transform $F(x) = \mathbf{U}^{\top}x$ to be the Fourier transform of graph signal, and $F'(x) = \mathbf{U}x$ to be reverse Fourier transform. This means, given any graph function $x \in \mathbb{R}^{N\times d}$, we can decompose $x$ in $N$ different frequency components:
\begin{align}
x=\sum_{i=1}^{N} \mathbf{U_{i}}(\mathbf{U_{i}}^{\top}x),
\label{eq:freq4}
\end{align}
where $\mathbf{U_{i}} \in \mathbb{R}^{N \times 1}$ is the $i$th column vector of $\mathbf{U}$. $d$ is the dimension of the graph signal on each vertex. $\mathbf{U_{i}}^{\top}x$ is the frequency component of $x$ on the $i$th frequency base. 

\begin{figure}[t]
  \centering
  \includegraphics[width=1\linewidth]{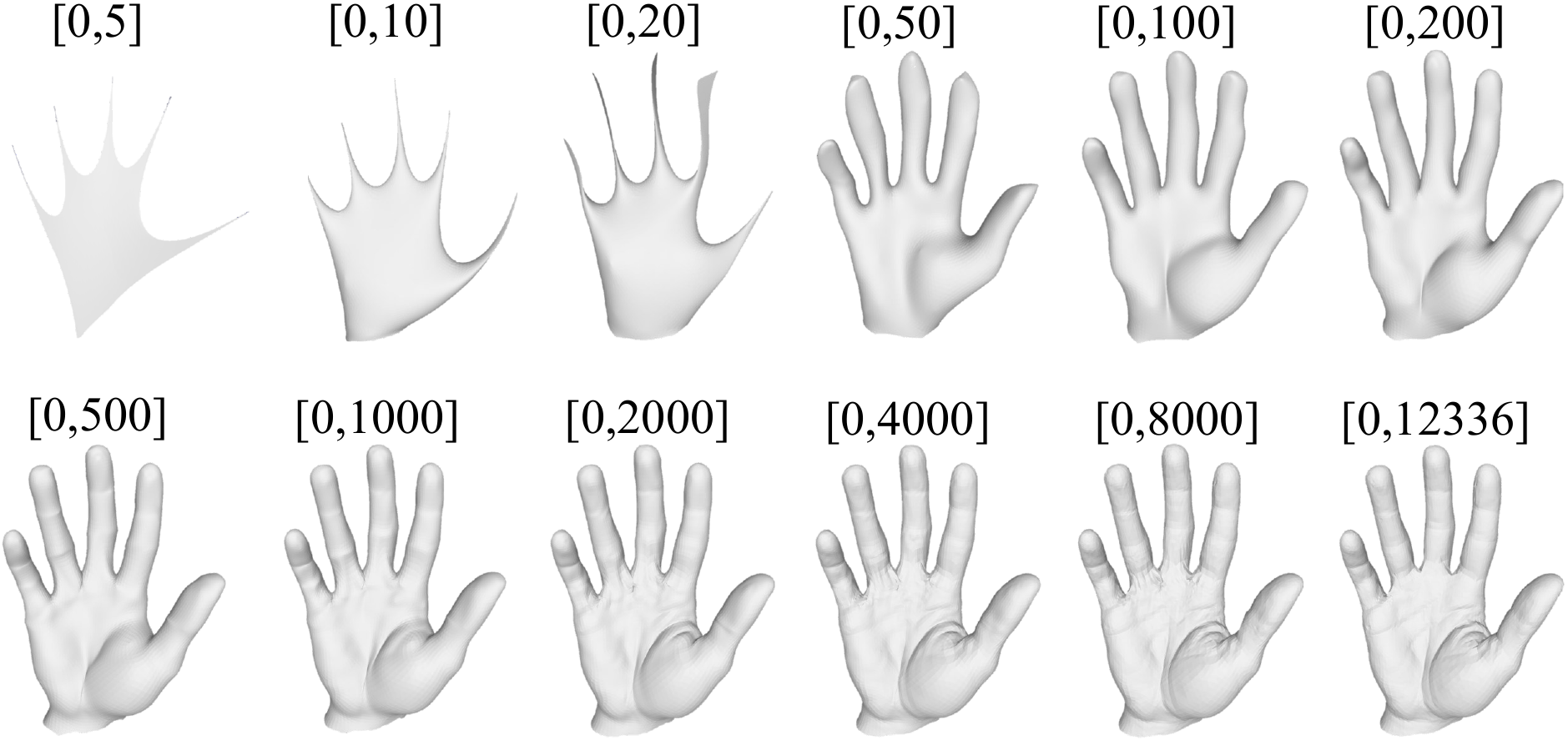}
  \caption{Frequency decomposition of 3D hand mesh. Cumulative frequency components start from frequency 0. The range shows the frequency band. For example, [0,20]
  means the signal of the first 21 frequencies (lowest 21) added together. We can see how the mesh shape changes when we gradually add higher frequency signals to the hand mesh. In general, the hand details increase as higher frequency signals are included.}
  \label{fig:12hands}
\end{figure}

Having \cref{eq:freq4}, we can decompose a hand mesh into frequency components. \cref{fig:spetrum} shows an example of a groundtruth mesh and its frequency decomposition result. The x-axis is the frequencies from low to high. The y-axis is the amplitude of each component in the logarithm. It is easy to observe that the signal amplitude generally decreases as the frequency increases. \cref{fig:12hands} shows the cumulative frequency components starting from frequency 0. We can see how the mesh shape changes when we gradually add higher frequency signals to the hand mesh. In general, the hand details increase as higher frequency signals are gradually included.

\subsection{Frequency Decomposition Loss}
\label{sec:loss}
\textbf{Frequency decomposition loss}. Conventional joint and vertex loss, such as the widely used pre-joint error loss \cite{chen2021model,hasson2020leveraging,baek2019pushing,boukhayma20193d,yang2020seqhand,zimmermann2017learning,guo2017towards,yang2019aligning,rudnev2021eventhands} and mesh pre-vertex error loss \cite{sengupta2020synthetic,pavlakos2018learning,yang2016estimation,kocabas2020vibe} commonly used in human body reconstruction, and Chamfer Distance Loss~\cite{jiang2018gal,wu2021density,athitsos2003estimating,mescheder2019occupancy} commonly used in object reconstruction and 3D point cloud estimation, all measure the error in the spatial domain. In that case, the signals of different frequency components are aliased together. As shown in \cref{fig:spetrum}, the amplitudes of low-frequency signals of hand shape are much larger than high-frequency signals, so when alias happens, the high-frequency signals will get overwhelmed, which means direct supervision on the spatial domain would mainly focus on low-frequency signals. Thus, spatial loss mostly does not drive the network to generate high-frequency details. Our experiments in \cref{sec:abla} also demonstrate this. 

To generate detailed information without being overwhelmed by low-frequency signals, we designed a loss function in the frequency domain. Specifically, we use graph frequency decomposition (\cref{sec:freq}) to define our frequency decomposition loss as
\begin{align}
    L_{F} = \frac{1}{F}\sum_{f=1}^{F}\log(\frac{\left\|\mathbf{U}_f^{\top}\hat{V}-\mathbf{U}_f^{\top}V_{gt}\right\|^2}{\left\|\mathbf{U}_f^{\top}\hat{V}\right\|\left\|\mathbf{U}_f^{\top}V_{gt}\right\| + \epsilon} + 1),
    \label{eq:loss_f}
\end{align}
where $F=N$ is the number of total frequency components, $\mathbf{U}_f$ is the $f$th frequency base, $\|\cdot\|$ is L2 norm, $\epsilon = 1 \times 10^{-8}$ is a small number to avoid division-by-zero, $\hat{V} \in \mathbb{R}^{N \times 3}$ and $V_{gt} \in \mathbb{R}^{N \times 3}$ are the predicted and groundtruth vertex locations, respectively. During training, for every frequency component, our loss reduces the influence of the amplitude of each frequency component, so that information on different frequency components would have equivalent attention. In \cref{tab:abla}, we demonstrate the effectiveness of the frequency decomposition loss.

\textbf{Total loss function.} We define the total loss function as:
\begin{align}
    L = \lambda_{J}L_{J} + \sum_{l=1}^{3}\left[ \lambda_{v}^{(l)}L_{v}^{(l)} + \lambda_{F}^{(l)}L_{F}^{(l)}\right],
    \label{eq:loss1}
\end{align}
where $l$ is the resolution level. $l=1$ is the lowest-resolution level and $l=3$ is the highest resolution level.
$L_{J}^{(l)}$ is 3D joint location error, $L_{v}^{(l)}$ is per vertex error, and $L_{F}^{(l)}$ is the frequency decomposition loss. $\lambda_{J}^{(l)}$, $\lambda_{v}^{(l)}$, and $\lambda_{F}^{(l)}$ are hyper-parameters. For simplicity, we refer $L_{J}^{(l)}$, $L_{v}^{(l)}$, and $L_{F}^{(l)}$ as $L_{J}$, $L_{v}$, and $L_{F}$ for the rest of the paper.

Following previous work \cite{chen2021s2hand,yang2016estimation}, we define 3D joint location error and per vertex loss as:
\begin{equation}
    L_{J} = \frac{1}{N_{J}}\sum_{j=1}^{N_{J}} \|\hat{J_j}-J_{gt,j}\|, \\
    L_{v} = \frac{1}{N}\sum_{i=1}^{N} \|\hat{v}_i-v_{gt,i}\|,
    \label{eq:loss2}
\end{equation}
where $\hat{J}_{j}$ and $J_{gt,j}$ are the output joint location and groundtruth joint location. $N_{J}$ is the number of joints. $\hat{v}_i$ and $v_{gt,i}$ are the estimated and groundtruth location of the $i$th vertex, and $N$ is the number of vertices.

\begin{figure*}[t]
  \centering
  \includegraphics[width=0.90\linewidth]{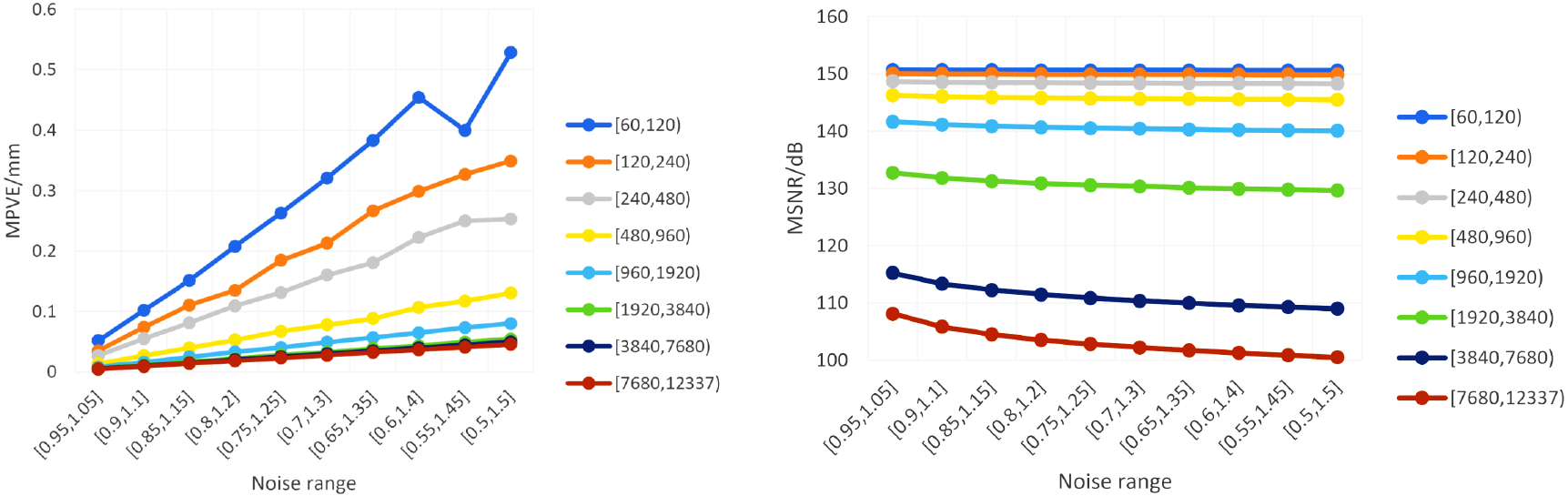}
  \caption{Evaluations using Euclidean distance and MSNR under different noise amplitudes in every frequency band. Each line of different color indicates a frequency band. The maximum and minimum frequencies are shown in the legend. On each line, every dot means adding a random amplitude noise to the mesh. The noise amplitude of each dot is evenly distributed in the ranges shown on the x-axis. The result validates that Euclidean distance is more sensitive to error in low-frequency bands, and MSNR is more sensitive to noise in high-frequency bands. Thus, compared to Euclidean distance, MSNR can better measure the error in high-frequency details.}
  \label{fig:metric1}
\end{figure*}

\begin{figure}[t]
  \centering
  \includegraphics[width=1\linewidth]{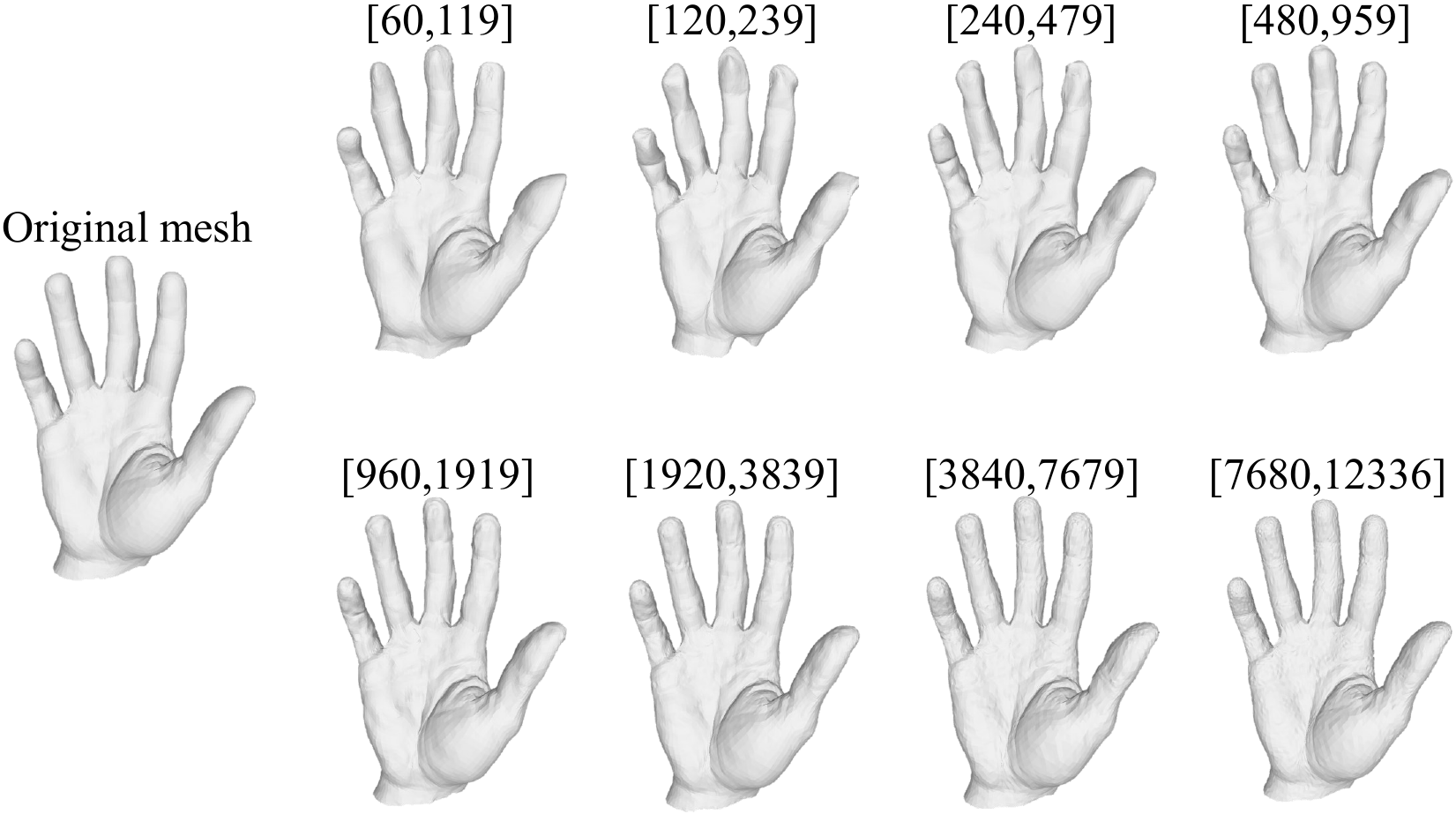}
  \caption{We show examples of Noisy Meshes. The meshes from left to right are meshes with a noise maximum amplitude of 0.6 and the frequency band changed from [60,119] to [7680,12336]. For easier visualization, we visualize the vertices location changes 5 times larger.}
  \label{fig:noisy}
\end{figure}

\section{Experiments}
\subsection{Datasets}
Our task requires detailed hand meshes for supervision. Because of the difficulty of acquiring 3D scan data, this supervision is expensive and hard to obtain in a large scale. One alternative plan is to generate meshes from multiview RGB images using multiview stereo methods. Considering the easy access, we stick to this plan and use the generated mesh as groundtruth in our experiments. We do all our experiments on the InterHand2.6M dataset \cite{Moon_2020_ECCV_InterHand2.6M}, which is a dataset consisting of multiview images, rich poses, and human hand pose annotations. The dataset typically provides 40-100 views for every frame of a hand video. Such a large amount of multiview information would help with more accurate mesh annotation. Finally, we remesh the result hand mesh into the same topology with our 3-level hand mesh template, respectively, so that we can provide mesh supervision for all 3 levels of our network. We use the resulting mesh as groundtruth for training and testing. In this paper, we use the mesh results provided in \cite{Moon_2020_ECCV_DeepHandMesh}, which are generated using multiview methods of \cite{Galliani_2015_ICCV}, and only use a subset of InterHand2.6m, due to the large number of data in the original dataset. The remeshing method and more dataset details can be found in supplementary material Section \Rmnum{4}. In \cref{fig:vis} (last column, ``groundtruth"), we show a few examples of the generated groundtruth meshes. Although these meshes are not the exact same as real hands, it is vivid and provides rich and high-fidelity details of human hands. This 3D mesh annotation method is not only enough to support our solution and verify our methods, but is also budget-friendly.

\subsection{Implementation Details.}  
We follow the network architecture in \cite{chen2021s2hand} to generate intermediate MANO results. We use EfficientNet \cite{tan2019efficientnet} as a backbone. The low-level, mid-level, and high-level features are extracted after the 1st, 3rd, and 7th blocks of EfficientNet, respectively. For each image feature, we use $1 \times 1$ convolutions to deduce dimensions. The channel numbers of $1 \times 1$ convolution are 32, 32, and 64 from low-level to high-level, respectively. After that, we project the initial human hand vertices to the feature maps, and sample a feature vector for every vertex using bilinear interpolation. The GCN graph has 778, 3093, and 12337 vertices at each resolution level.

In the training process, we first train \cite{chen2021s2hand} network, and then use the pretrained result to train our scalable network. For training \cite{chen2021s2hand}, we use their default hyper-parameters, set the learning rate to $1 \times 10 ^{-4}$, and set batch size to 48. When training GCN network, we set $\lambda_{J}$ to be 1, set $\lambda_{v}^{(1)}$ and $\lambda_{F}^{(1)}$ to be 1 and 60, set $\lambda_{v}^{(2)}$ and $\lambda_{F}^{(2)}$ to be also 1 and 60, and set $\lambda_{v}^{(3)}$ and $\lambda_{F}^{(3)}$ to be 1 and 100. The learning rate is set to $5 \times 10 ^{-4}$ for GCN and 1e-4 for the rest of the network. The batch size is set to 28. The training process takes about 25 hours on 1 NVIDIA GTX3090Ti GPU for 150 epochs. In reference, we use a smooth kernel to post-process the mesh to reduce sharp changes.
More details of post-processing will be found in Supplementary Materials Section \Rmnum{3}.

\subsection{Quantitative Evaluation}
\label{sec:metric}
We use mean per joint position error (MPJPE) and Chamfer distance (CD) to evaluate the hand pose and coarse shape. Besides, to better evaluate personalized details, we also evaluate our mesh results using the proposed mean signal-to-noise ratio (MSNR) metric. 

\setlength{\tabcolsep}{4pt}
\begin{table}
\begin{center}
\resizebox{0.8\linewidth}{!}{%
\begin{tabular}{llll}
\hline\noalign{\smallskip}
Method & MPJPE/mm $\downarrow$ & CD/mm $\downarrow$ & MSNR $\uparrow$\\
\noalign{\smallskip}
\hline
\noalign{\smallskip}
    MANO &13.41 &6.20 &-2.64 \\
    Ours-level 1 &13.25 & 5.53 & -2.70\\
    Ours-level 2 &13.25 & 5.49 & -2.62 \\
    Ours-level 3 &\textbf{13.25} & \textbf{5.49} & \textbf{-0.68} \\
\hline
\end{tabular}%
}
\caption{Results on InterHand2.6M dataset. For MPJPE and CD, lower is better. For MSNR, higher is better. 
As shown in the table, the proposed method improves the accuracy of hand surface details. While our method generates better shape details in a scalable manner, the joint locations and the overall shape also become slightly more accurate.}
\label{tab:sota}
\end{center}
\end{table}

\setlength{\tabcolsep}{4pt}
\begin{table}
\centering
\resizebox{0.85\linewidth}{!}{%
  
  \begin{tabular}{lllll}
  \hline\noalign{\smallskip}
    Level & \#parameter & GFLOPS & \#vertices & \#faces \\
   \noalign{\smallskip}
    \hline
    \noalign{\smallskip}
    baseline & 14.5M & 1.8& 778& 1538\\
    1 & 14.5M & 1.9& 778& 1538\\
    2 & 14.5M& 2.5& 3093& 6152\\
    3 & 14.7M& 4.8& 12337& 24608\\
    \hline
    \end{tabular}%
}
\caption{The mesh size and the resources needed for generating different resolution levels of meshes.}
\label{tab:scale}
\end{table}

\textbf{Mean Signal-to-Noise Ratio (MSNR).}
Previous metrics for 3D hand mesh mostly calculate the Euclidean distance between the results and the groundtruth. Although in most cases, Euclidean distance can roughly indicate the accuracy of the reconstruction results, it is not consistent with human cognitive standards: it is more sensitive to low-frequency errors, but does not perform well in personalized detail distinction or detailed shape similarity description.

Thus, we propose a metric that calculates the signal-to-noise ratio in every frequency base of the graph. We define our Mean Signal-to-Noise Ratio (MSNR) metric as
\begin{equation}
    \mathrm{MSNR} 
    =\frac{1}{F}\sum_{f=1}^{F} \log(\frac{\left\|\mathbf{U}_f^{\top}\hat{V}\right\|}{\left\|\mathbf{U}_f^{\top}\hat{V} - \mathbf{U}_f^{\top}V_{gt}\right\| + \epsilon}),
\end{equation}
where $F=N$ is the total number of frequency components and $S_{f}$ is the signal-to-noise ratio of the $f$th frequency component. $\mathbf{U}_f$, $\hat{V}$, and $V_{gt}$ have the same meaning as in \cref{eq:loss_f}. $\epsilon=1 \times 10 ^{-8}$ is a small number to avoid division-by-zero. Thus, the maximum of $S_{f}$ is $8$. By this design, the SNR of different frequency components would not influence each other, so we can better evaluate the high-frequency information compared to the conventional Euclidean Distance.

We designed an experiment on InterHand2.6m to validate the effectiveness of our metric in evaluating high-frequency details. We add errors of 8 different frequency bands to the hand mesh. For each frequency band, the error amplitude is set under 10 different uniform distributions. As shown in \cref{fig:metric1}, 
we measure the MPVE and MSNR for every noise distribution on every frequency band, to see how the measured results of the two metrics change with the noise amplitude in each frequency band. The result shows that in the low-frequency part, MPVE increases fast when the noise amplitude increases (the upper lines), but in high-frequency bands, the measured result changes very slowly when the noise amplitude increases. MSNR behaves completely differently from MPVE. It is more sensitive to noise in the high-frequency band than in the low-frequency band. Thus, compared to Euclidean distance, MSNR better measures the error in high-frequency details. \cref{fig:noisy} shows a few examples of noisy meshes.

\begin{figure}[t]
  \centering
  \includegraphics[width=1\linewidth]{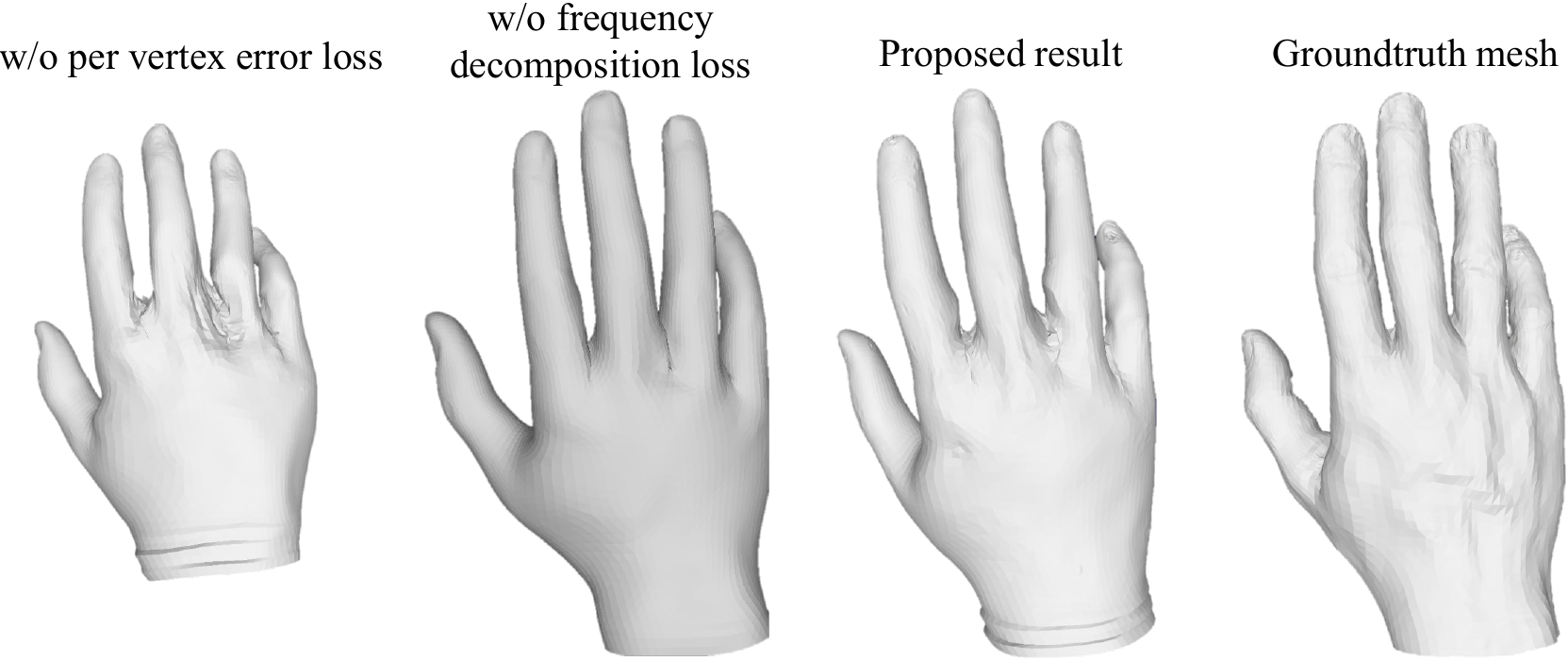}
  \caption{Visualization results of “w/o frequency decomposition loss” and "w/o per vertex error loss" in \cref{sec:abla}. As shown, if we do not use frequency decomposition loss, the mesh result we get tends to be smoother with less personalized details. If we do not use per-vertex error loss, the mesh's low-frequency information is not well-learned. The mesh we generate will have an overall shape deformation.}
  \label{fig:sup6}
\end{figure}

\textbf{Evaluation on InterHand2.6M dataset.} We report mean per joint position error (MPJPE), Chamfer distance (CD), and mean signal-to-noise ratio (MSNR) to evaluate the overall accuracy of reconstructed hand meshes. \cref{tab:sota} shows the comparison among 3 levels of our proposed method and MANO. As shown in the table, the proposed method improves the accuracy of hand surface details by a large margin (as indicated by MSNR). We also observe that, while our method generates better shape details in a scalable manner, the joint locations and the overall shape of the output meshes also become slightly more accurate (as indicated by MPJPE and CD). Here, the MSNR of MANO, Ours-level 1, and Ours-level 2 are calculated after subdividing their meshes into the same resolution as Ours-level 3.

\setlength{\tabcolsep}{4pt}

\begin{table}
\begin{center}
\resizebox{1\linewidth}{!}{%

  \begin{tabular}{llll}
  \hline\noalign{\smallskip}
    Method & MPJPE/mm $\downarrow$ & CD/mm $\downarrow$ & MSNR $\uparrow$\\
   \noalign{\smallskip}
    \hline
    \noalign{\smallskip}
    proposed &\textbf{13.25} & \textbf{5.49}& \textbf{-0.68}\\
    \hline
    w/o skip connected feature &14.20& 5.85& -0.70\\
    w/ average pooling feature &13.95& 5.59& -1.10\\
    \hline
    w/o frequency decomposition loss &14.50 &5.86 &-1.80\\
    w/o per vertex error loss &14.24& 67.8& --0.87\\
   
    \hline
\end{tabular}
}
\label{tab:abla}
\caption{Ablation study on the feature skip connection design and the effect of loss functions. From the result, we can see that the frequency decomposition loss helps learn mesh details and the per-vertex error loss helps constrain the overall shape.}
\end{center}
\end{table}

\begin{figure*}[t]
  \centering
  \includegraphics[width=0.90\linewidth]{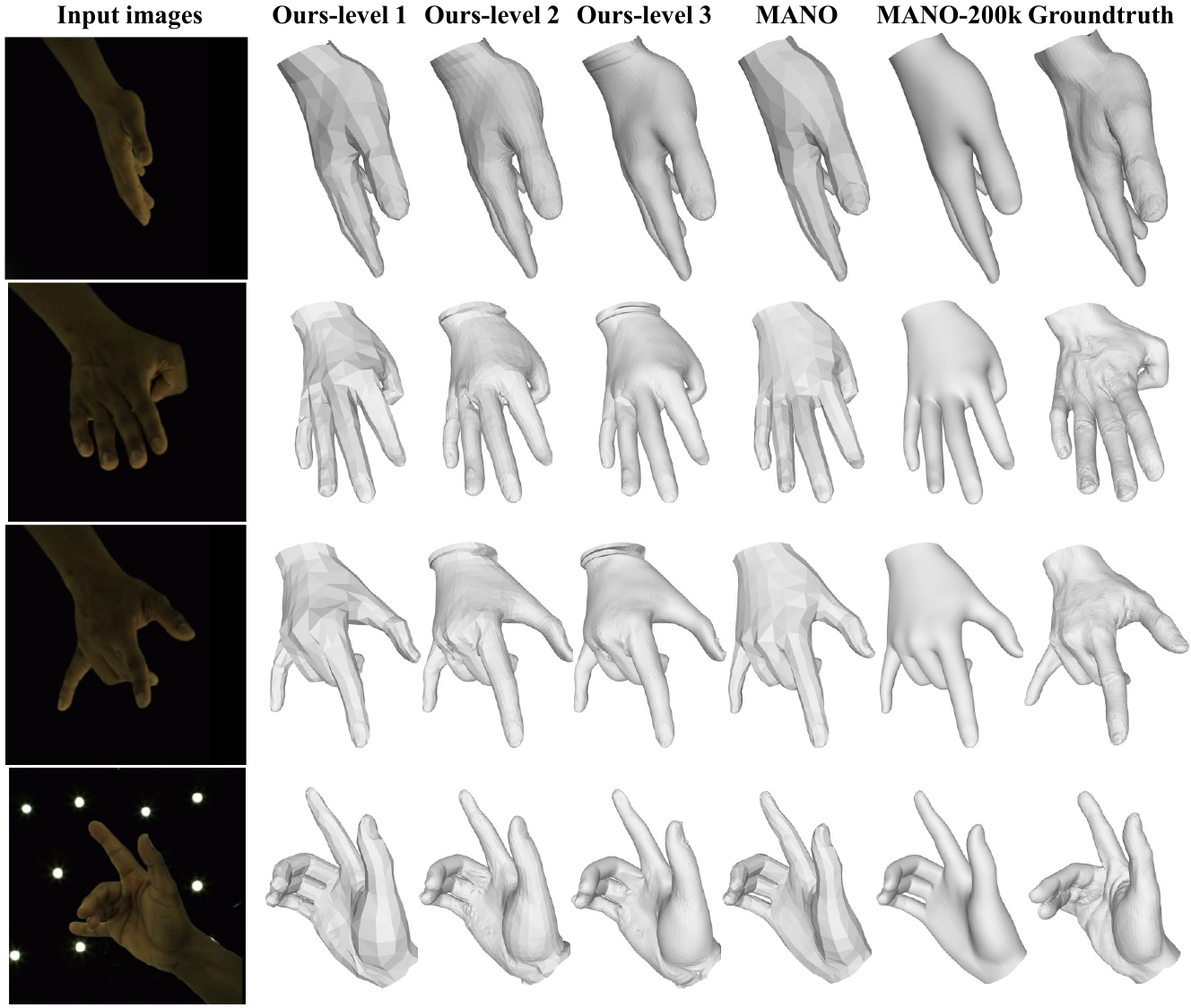}
  \caption{Qualitative reconstruction results. (Best viewed in magnification.) The columns from left to right are input images, our level 1-3 output mesh, MANO mesh, MANO mesh subdivided to 12.3k vertices (same vertex number as our mesh), and groundtruth, respectively. We can see that even if we upsample MANO into the same number of vertices as our mesh, it still does not provide comparable personalized details as our results.
  }
  \label{fig:vis}
\end{figure*}

\subsection{Ablation Study}
\label{sec:abla}

We conduct several experiments to demonstrate the effectiveness of the feature skip connection design (in \cref{fig:pipeline}).
and different loss functions. The results are shown in \cref{tab:abla}. From the result, we observe that our projection-to-feature-map skip connection design leads to performance improvement in all three metrics.
For the loss functions, we observe MSNR degrades when the frequency decomposition loss is removed, indicating inferior mesh details.
Removing the per-vertex error loss dramatically increases the Chamfer distance, indicating that the overall shape is not well constrained.
The visualization results of the latter 2 experiments are shown in \cref{fig:sup6}, if we do not use frequency decomposition loss, the mesh result we get tends to be smoother with less personalized details. If we do not use per-vertex error loss, the mesh's low-frequency information is not well-learned. The mesh we generate will have an overall shape deformation.

\textbf{Scalable design.} We also demonstrate the scalable design of the proposed network by analyzing the resource needed at each resolution level (\cref{tab:scale}). In general, higher resolution levels require more computational resources in the network, and more resources to store and render the mesh. Still, our approach supports scalable reconstruction and can be applied to scenarios with limited computational resources.
Here, ``baseline" means only generating the MANO mesh in our network.

\textbf{Visualization Results.}
The qualitative reconstruction results are shown in \cref{fig:vis}. We observe that even when MANO is upsampled to 200k
vertices, it still does not capture personalized details while our results provide better shape details. 
More qualitative results can be found in the Supplementary Material Section \Rmnum{5}.

\section{Conclusion}
We provided a solution to reconstruct high-fidelity hand mesh from monocular RGB inputs in a scalable manner. We represent the hand mesh as a graph and design a scalable frequency split network to generate hand mesh from different frequency bands. To train the network, we propose a frequency decomposition loss to supervise each frequency component. Finally, we introduce a new evaluation metric named Mean Signal-to-Noise Ratio (MSNR) to measure the signal-to-noise ratio of each mesh frequency component, which can better measure the details of 3D shapes. The evaluations on benchmark datasets validate the effectiveness of our proposed method and the evaluation metric in terms of recovering 3D hand shape details. 

\section*{Acknowledgments}
This work is supported in part by a gift grant from OPPO.

\clearpage
{\small
\bibliographystyle{ieee_fullname}
\bibliography{egbib}
}

\appendix

\def\model{SAD}

\section{Detailed Network Architecture} 
\begin{figure*}[t]
  \centering
  \includegraphics[width=1\linewidth]{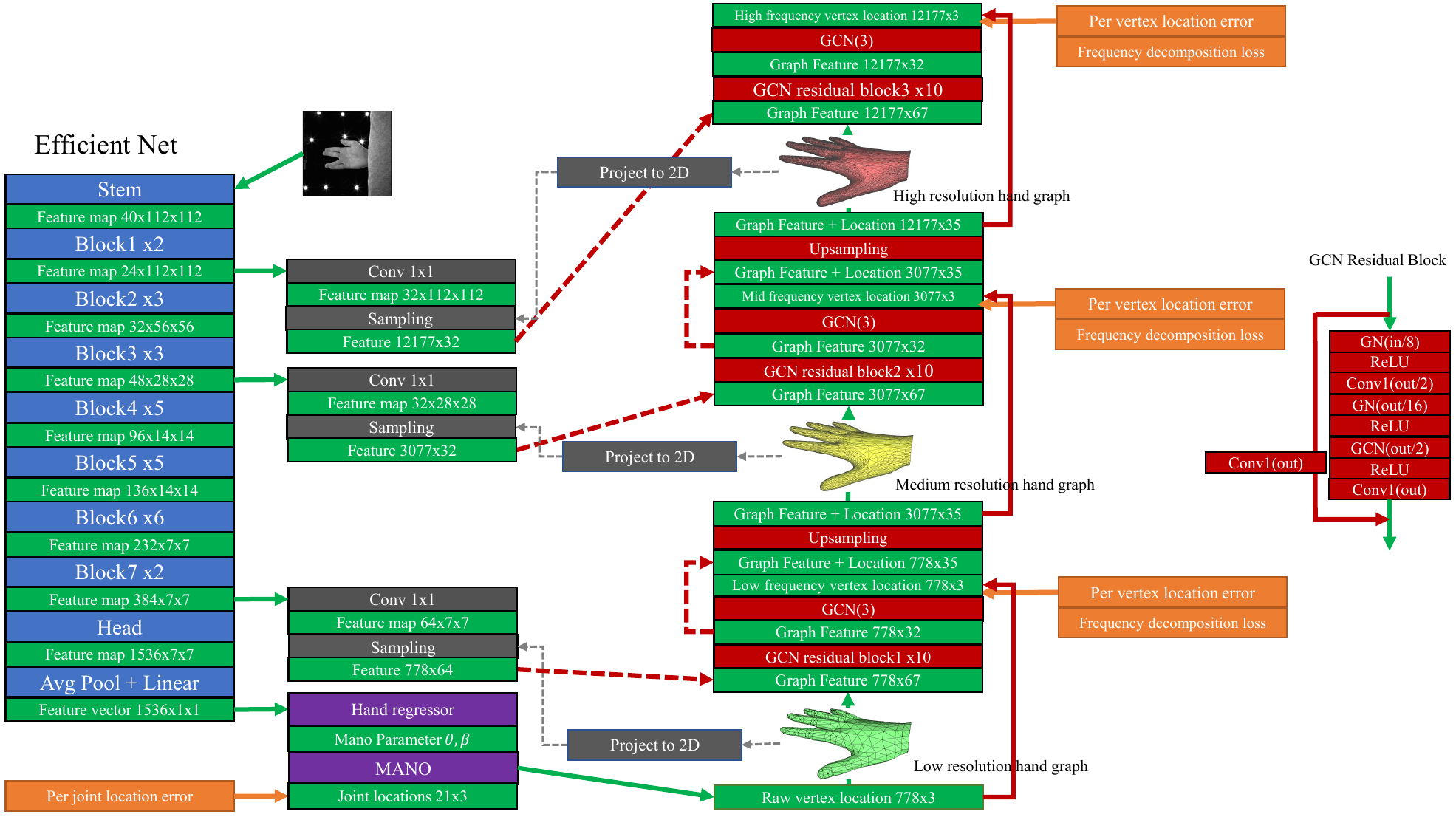}
  \caption{The detailed network architecture.}
  \label{fig:sup1}
\end{figure*}

We proposed a detailed network architecture of our approach in \cref{fig:sup1}. The green boxes are the features, in which we note the feature dimensions. The blue boxes represent blocks of EfficientNet \cite{tan2019efficientnet}. The red boxes represent GCN blocks. The GCN residual blocks in the network are designed following the manner of \cite{kolotouros2019convolutional}. Details of the residual blocks are shown on the right of the figure. The gray boxes are the feature skip-connection part. To get multi-level image features from feature maps, we project the vertices into the feature maps, and use a bilinear interpolation technique to sample features. 
We will illustrate the process more in \cref{sec:sec2}. 
The purple boxes are the sub-network used to generate MANO mesh. The orange boxes indicate the annotation we used. The green arrows are feature streams and the red lines are skip connections.

We fetch skip-connected features from the output of EfficientNet Block 1, Block 3, and Block 7. The features are used as parts of the input of the GCN. The GCN has 3 levels. At each level, the input features go through a 10-layer GCN Residual Block, then output a feature vector and a 3D location at each vertex. The 3D locations are used as intermediate output and for supervision. The features are used as a part of the input for the next level. At the third level, we only output the 3D location of each vertex as the final mesh.

\begin{figure}[t]
  \centering
  \includegraphics[width=0.7\linewidth]{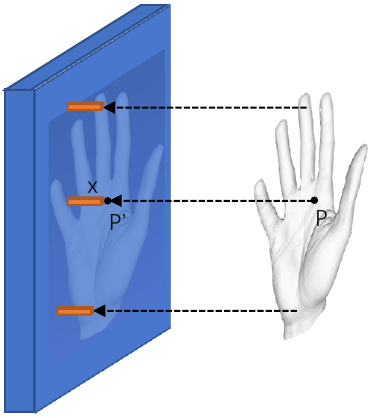}
  \caption{Skip-connected Feature Sampling.}
  \label{fig:sup2}
\end{figure}

\section{Skip-connected Feature Sampling}  
\label{sec:sec2}
In \cref{fig:sup1}, the features fetched from EfficientNet are feature maps. We want to transfer them into feature vectors and put them on the vertices without losing spatial information. Thus, we design a feature sampling strategy to put the local image feature on each graph vertex. As shown in \cref{fig:sup2}, we use orthodox projection to find the feature vector for each vertex on the feature map. For every vertex $P$, we calculate the projection point $P^{\prime}$ on the feature map. Then, we extract the feature vector $x \in \mathbf{R}^{c}$ using bilinear interpolation at point $P^{\prime}$, where $c$ is the feature map channel number. The total output feature dimension is $N \times c$, where $N$ is the number of graph vertices.

\section{Mesh Post-processing}  
\label{sec:sec4}

\begin{figure}[t]
  \centering
  \includegraphics[width=1\linewidth]{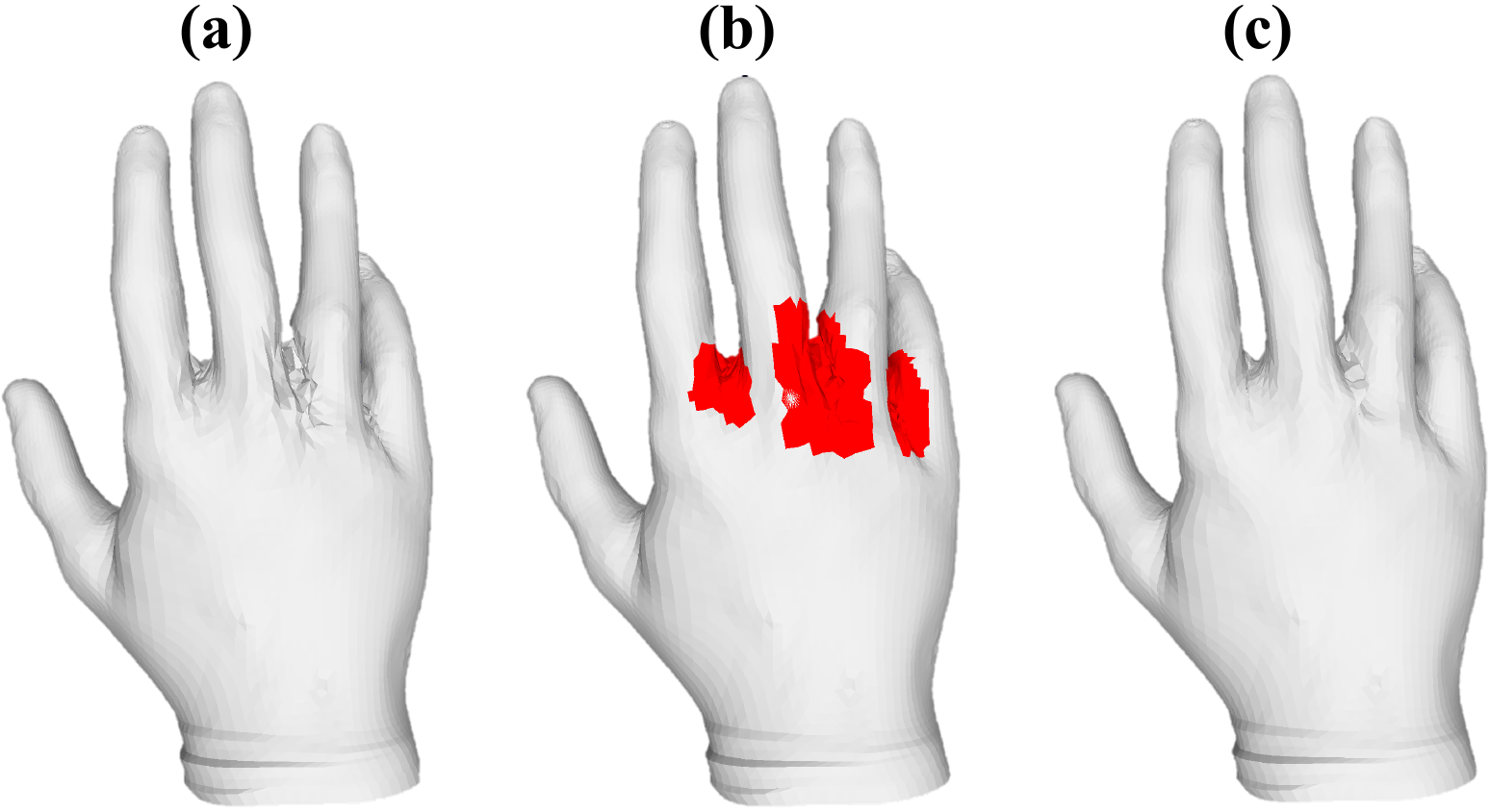}
  \caption{Mesh pose-processing. a) Original mesh. b) Smoothing mask (Red). c) Final result.}
  \label{fig:sup3}
\end{figure}

We do a post-process on the third-level mesh. Due to the flaws of groundtruth mesh (shown in \cref{fig:sup4}), some of our output mesh also have similar structure flaws. To tackle this problem, we designed a smooth mask to reduce the flaws. \cref{fig:sup3} shows the output of the network, our smooth mask, and our final mesh result. As we can see, the flaws are highly reduced. Note that, this flaw is caused by the noisy groundtruth, so it can also be reduced by a better remeshing of the training data in the future.

\begin{figure}[t]
  \centering
  \includegraphics[width=0.8\linewidth]{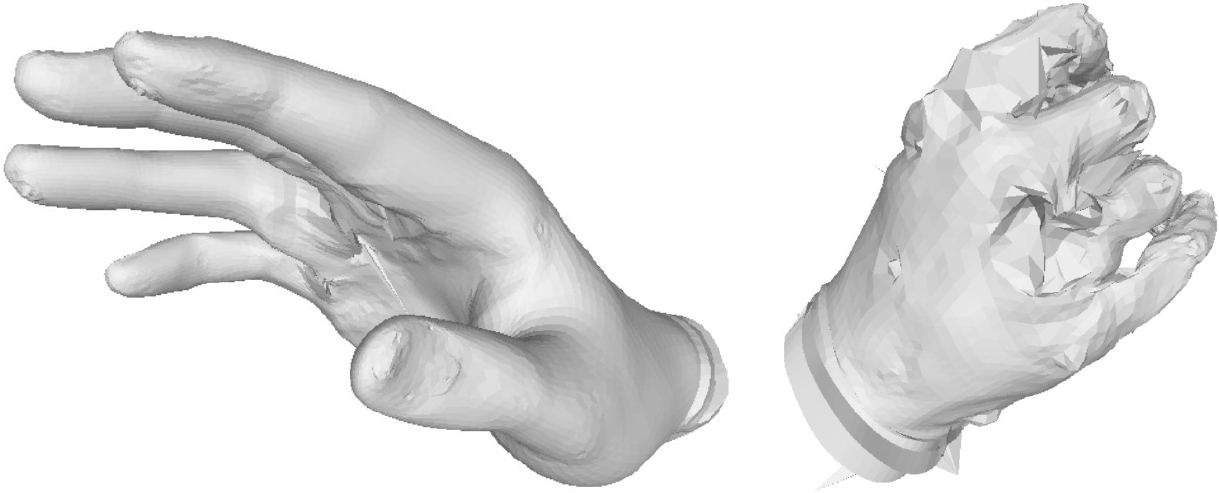}
  \caption{Failure cases.}
  \label{fig:failure}
\end{figure}

\section{Remeshing Procedure}  
\label{sec:sec5}

We try to use the multiview stereo (MVS) generated mesh provided in \cite{Moon_2020_ECCV_DeepHandMesh}. However, the MVS mesh has about 500k vertices on each mesh. The large vertex number mesh with high redundancy makes our training process much slower. Moreover, without a fixed topology, the choices of shape supervision are limited. For example, we would not be able to use the per vertex loss and frequency decomposition loss for training.

\begin{figure*}[t]
  \centering
  \includegraphics[width=0.9\linewidth]{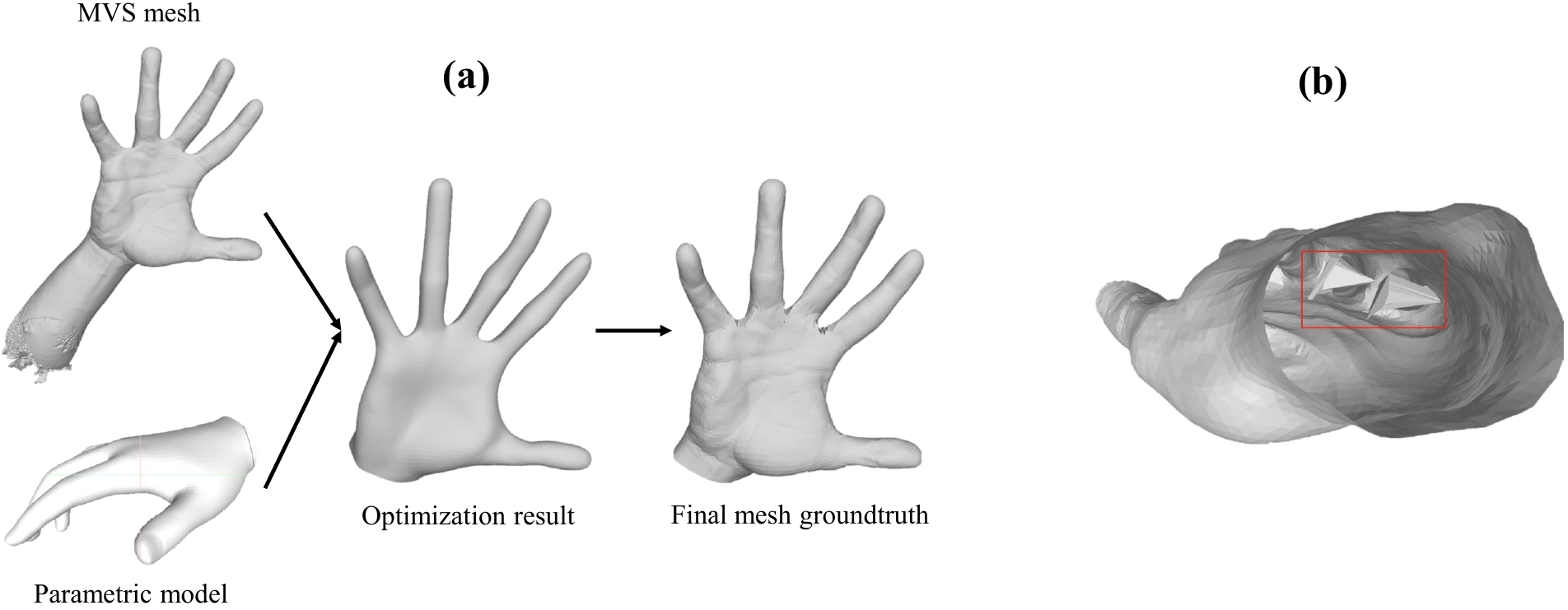}
  \caption{a) Remeshing procedure. b) Example of groundtruth flaws}
  \label{fig:sup4}
\end{figure*}

Thus, we designed a remeshing technic to transfer the mesh generated in the multiview stereo (MVS) method into a unified topology. The algorithm is shown in \cref{fig:sup4}\textcolor{red}{a}. First, we align the MVS mesh with a parametric template mesh. Here, we use template meshes designed in the main paper Section 3.2.
Second, we use an optimization approach to calculate a set of pose and shape parameters, so that the template mesh becomes a coarse approximation of the MVS mesh. Finally, we use the closet point on the MVS mesh as a substitute for each vertex on the parametric mesh. This procedure would preserve the detailed shape and the topology of the parametric template at the same time. In our experiments, we generate 3 resolution levels of groundtruth mesh for supervision, and use the third level for testing.

However, despite the good attributes of the groundtruth meshes, some of them still have flaws. \cref{fig:sup4}\textcolor{red}{b} shows an example of the mesh flaws inside the mesh (red rectangle). It happens because some of the vertices on the parametric mesh find the wrong corresponding vertices on the MVS mesh. These groundtruth mesh flaws will eventually cause defects in generated mesh (shown in \cref{fig:sup3}). We have largely reduced the flaws of our mesh using the mesh post-processing method mentioned in \cref{sec:sec4}.

\section{More Visualization Results}  
\begin{figure*}[t]
  \centering
  \includegraphics[width=1\linewidth]{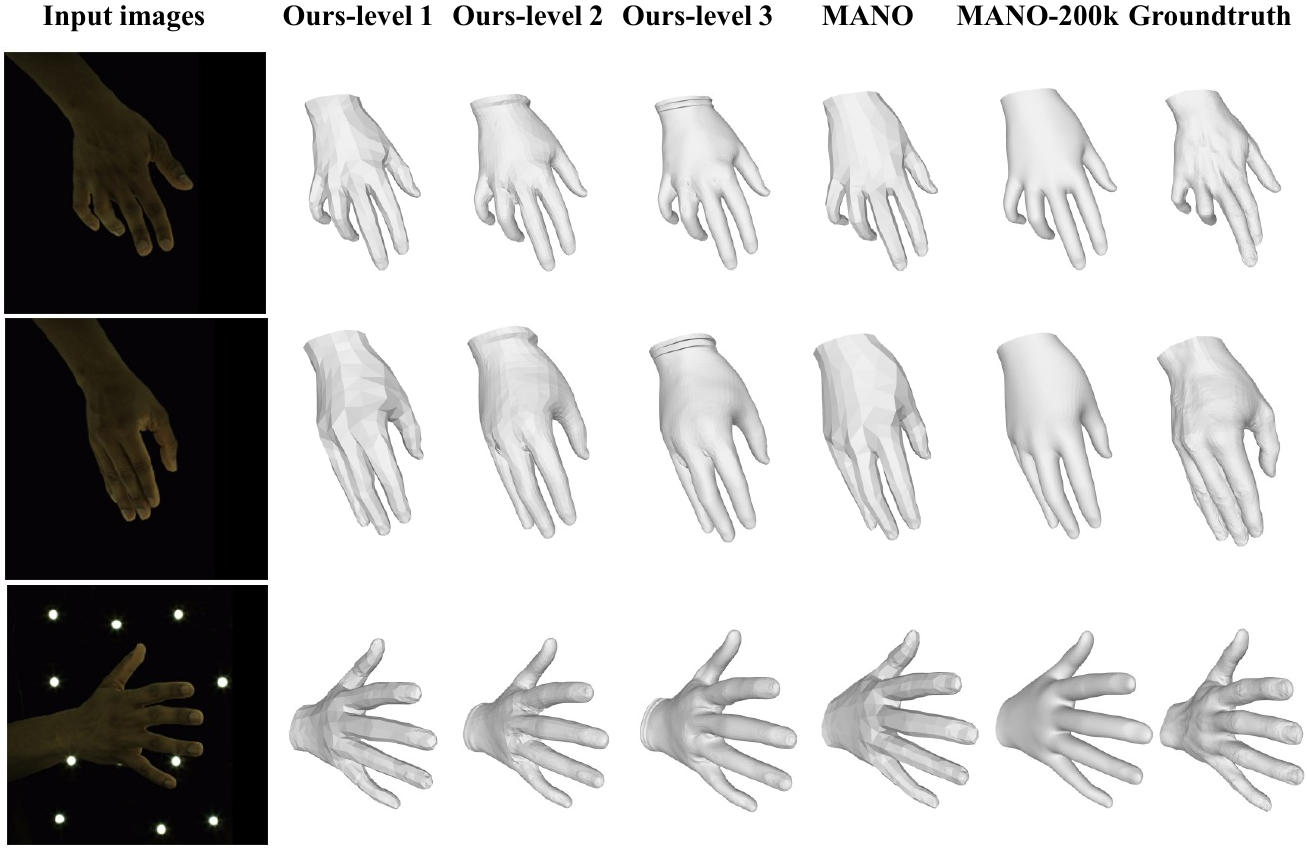}
  \caption{More visualization results.}
  \label{fig:sup7}
\end{figure*}

We show more visualization results of our proposed method in \cref{fig:sup7}.

\section{Failure Cases}
We show in \cref{fig:failure} a few failure cases where our method generates hand meshes with flaws. Most of these flaws are caused by groundtruth flaws in remeshing (shown in \cref{fig:sup4}\textcolor{red}{b}). 

\section{Future Works and Discussions}
In future works, our backbone can be replaced with more recent work such as those in \cite{zhang2022morphmlp, pan20232d, lou2023cfpnet, lou2023caranet, zhai2020two, zhang2023neural11, zh2023toward}. The object detection and segmentation-related network can be helpful for hand-related tasks. We would also improve the remeshing procedure to reduce the artifacts. Besides, we would also improve our method to tackle the in-the-wild hand reconstruction problem. Moreover, the frequency decomposition approach can be easily expanded to improve the details of human body reconstruction works such as \cite{luan2021pc, zhang2021learning, wang2022best, gong2022self, gong2023progressive}.

\end{document}